\documentclass[11pt]{article}

\usepackage[margin=1in]{geometry}
\usepackage{times}
\usepackage[hyphens]{url}
\usepackage{graphicx}
\usepackage{booktabs}
\usepackage{multirow}
\usepackage{amsmath,amssymb,amsthm}
\usepackage{algorithm}
\usepackage{algorithmic}
\usepackage{xcolor}
\usepackage{microtype}
\usepackage{hyperref}
\usepackage{authblk}
\usepackage{natbib}

\usepackage{amsmath,amsfonts,bm}

\def\eqref#1{equation~\ref{#1}}

\def\1{\bm{1}}

\DeclareMathAlphabet{\mathsfit}{\encodingdefault}{\sfdefault}{m}{sl}
\SetMathAlphabet{\mathsfit}{bold}{\encodingdefault}{\sfdefault}{bx}{n}

\newcommand{\softmax}{\mathrm{softmax}}

\newcommand{\method}{\textsc{NSAI-Fuzz}}
\newcommand{\rcw}{\textsc{RCW}}
\newcommand{\csg}{\textsc{CSG}}
\newcommand{\ind}{\mathbb{1}}
\newcommand{\AUC}{\operatorname{AUC}}
\newcommand{\pp}{\,\mathrm{pp}}

\newtheorem{proposition}{Proposition}

\title{One Fuzzer, Many Reasoners: Adaptive Neuro-Symbolic Fuzzing with Reliability-Calibrated and Context-Gated Guidance}

\author[1]{Jinghao Wen}
\author[2]{Xianglong Yu}
\author[1]{Shaohuang Wang}
\author[1]{Xun Jiao}

\affil[1]{Department of Electrical and Computer Engineering, Villanova University, Villanova, PA, USA}
\affil[1]{School of Arts and Sciences, University of Pennsylvania, Philadelphia, PA, USA}

\date{}

\begin{document}
\maketitle

\begin{abstract}
Neural-symbolic AI (NSAI) combines learned perception with programs, constraints, and compositional reasoning, but remains difficult to test because small semantics-preserving changes can alter grounded symbols and trigger discontinuous reasoning failures. Existing neural fuzzers mainly rely on neural coverage or representation-space feedback and do not provide a common testing interface for heterogeneous NSAI systems. We introduce \method, the first unified semantics-aware fuzzing framework that separates task-specific valid mutation from architecture-neutral guided search and represents each candidate using five availability-aware, rank-normalized risk signals: boundary pressure, grounding uncertainty, representation drift, constraint violation, and archive novelty. We further propose Reliability-Calibrated Weighting (\rcw), which learns dataset--model-level signal allocations from calibrated reliability, and Contextual Signal Gating (\csg), which adapts signal weights to the current candidate pool. We evaluate ISED, DOLPHIN, NASR, and NeSyCoCo across four reasoning tasks, nine compatible dataset--model cells, 20 policies, and 1,800 on-policy runs. Fixed five-signal guidance outperforms Random and Uniform in every cell, with gains of up to $19.115\pp$ and $3.280\pp$, while \rcw\ and \csg\ further improve over Fixed in all nine cells by up to $6.621\pp$. Under matched mutation and query budgets, \rcw\ and \csg\ achieve 16.17\% and 16.06\% aggregate failure discovery, versus 6.06\% for the strongest neural-fuzzing baseline, while retaining substantially more deduplicated failure behaviors. Wall-clock evaluation further shows that TensorFuzz requires $1.11\times$ the runtime of \rcw\ while achieving only $0.45\times$ its failure throughput. These results show that effective and efficient fuzzing of heterogeneous NSAI systems benefits from a common neural--symbolic risk interface with adaptive signal allocation.
\end{abstract}

\section{Introduction}
Neural-symbolic AI (NSAI) combines learned perception with explicit reasoning.
A neural module may ground pixels into digits, objects, or
concepts; a downstream component then executes an arithmetic program, enforces
Sudoku constraints, or composes a language-conditioned concept program.
Representative systems include DeepProbLog
\cite{manhaeve2018deepproblog}, ISED \cite{solkobreslin2024ised}, DOLPHIN
\cite{naik2025dolphin}, NASR \cite{cornelio2023nasr}, and NeSyCoCo
\cite{kamali2024nesycoco}. This interface enables structured reasoning, but
also creates a sharp reliability boundary: a small change in neural evidence
can replace a grounded symbol, alter an execution path, or violate a global
constraint.

Testing such systems is challenging. Clean accuracy does not characterize how
failures are distributed around correctly solved inputs, while conventional
adversarial attacks are an imperfect fit because symbolic execution may be
non-differentiable or combinatorial. Moreover, a perturbation is meaningful
only if it preserves the task semantics: changing the answer of a Sudoku
instance or the referent of a grounding command does not constitute a valid
robustness failure. Existing neural fuzzers \cite{pei2017deepxplore,tian2018deeptest,xie2019deephunter,odena2019tensorfuzz} such as DeepHunter and TensorFuzz guide search using neuron coverage, activation-space novelty, or other
representation-specific feedback. These mechanisms remain useful for exploring
neural states, but they do not directly expose NSAI-specific risks such as
grounding instability, symbolic constraint violation, or execution-boundary
changes, nor do their original interfaces naturally span heterogeneous
neural--symbolic systems.

To the best of our knowledge, we present the first unified fuzzing framework
specifically designed to test heterogeneous NSAI systems. \method\ separates
\emph{what mutations are valid} from \emph{which valid candidates should be
explored}. Each task adapter generates semantics-preserving mutations and
provides an oracle that verifies that the expected answer is unchanged.
Search is then performed through a common architecture-neutral interface rather
than through model-specific coverage metrics.
The common interface represents each valid candidate using five signals:
boundary pressure, grounding uncertainty, representation drift, constraint
violation, and archive novelty. Pool-wise rank normalization removes scale
differences, while unavailable signals are explicitly masked, enabling a common
search procedure across heterogeneous neural--symbolic interfaces.

Our experiments first establish that structured multi-signal guidance is
effective within this common framework. Fixed five-signal guidance outperforms
Random and Uniform in all nine dataset--model cells, with gains of up to
$19.115\pp$ and $3.280\pp$, respectively. Single-signal and leave-one-out
diagnostics further show that signal importance varies substantially across
tasks and models. Thus, the main challenge is not whether to use the five
signals, but how to allocate their weights for each model--task setting.

We address this challenge with two adaptive mechanisms. Reliability-Calibrated
Weighting (\rcw) learns a cell-level allocation from signal reliability, while
Contextual Signal Gating (\csg) adapts the allocation to the current candidate
pool during search. We evaluate ISED, DOLPHIN, NASR, and NeSyCoCo across four
tasks, yielding nine compatible cells, 20 policies, ten paired attack seeds,
and 1,800 on-policy runs. Both adaptive methods outperform Fixed,
with gains of up to $6.621\pp$. We further compare \method\ against five
matched neural-fuzzing feedback policies, DeepHunter, TensorFuzz, DLFuzz,
NSGen-transfer, and CriticalFuzz, under identical inputs, mutation operators,
semantic constraints, frontier width, hardware, and query budgets. The strongest
aggregate neural baseline, TensorFuzz, reaches a $6.06\%$ failure discovery
rate, whereas \rcw\ and \csg\ achieve $16.17\%$ and $16.06\%$, corresponding
to $2.67\times$ and $2.65\times$ higher failure discovery, respectively.
After exact-output and activation-similarity deduplication, TensorFuzz retains
6,210 failure clusters, compared with 11,508 for \rcw\ and 12,058 for \csg, indicating
that the higher yield is not explained only by repeated near-duplicate
failures.

\begin{figure*}
	\centering
	\includegraphics[width=0.9\linewidth]{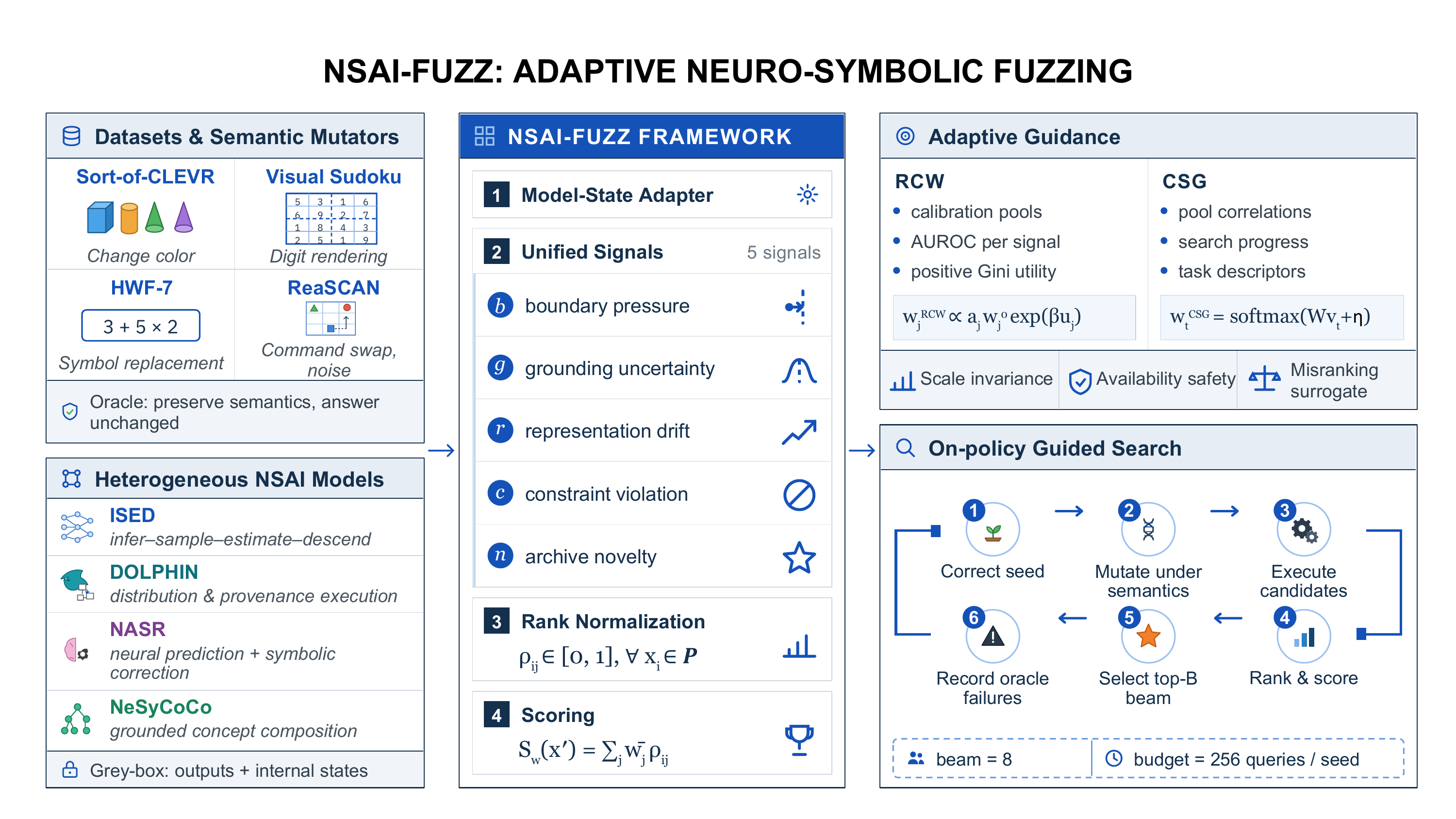}
	\caption{Overview of \method. Task-specific semantic mutators generate valid candidates,
which are mapped by a common adapter to five unified guidance signals.
\rcw\ and \csg\ adapt their weights to guide on-policy search across
heterogeneous NSAI systems.}
	\label{fig:framework}
\end{figure*}

We additionally evaluate wall-clock efficiency across both the principal
\method\ policies and matched neural-fuzzing baselines. Using \rcw\ as the
common normalization reference, Random requires only $0.71\times$ the online
runtime but achieves just $0.20\times$ the failure throughput, while
TensorFuzz requires $1.11\times$ the runtime of \rcw\ but delivers only
$0.45\times$ its throughput. \csg\ exhibits nearly identical online cost to
\rcw\ ($1.01\times$ runtime) and retains comparable failure throughput
($0.98\times$). These results show that the higher failure yield of adaptive
NSAI guidance is not obtained by spending substantially more wall-clock time.
Figure~\ref{fig:framework} summarizes \method, from semantics-preserving
mutation and unified signal extraction to adaptive weighting and on-policy
guided search. Our contributions are:
\begin{itemize}
    \item To the best of our knowledge, we present the first unified fuzzing
    framework specifically designed for heterogeneous neural-symbolic AI
    systems, separating task-specific semantics-preserving mutation from
    architecture-neutral guided search.

    \item We introduce five availability-aware, rank-normalized guidance
signals. Fixed guidance outperforms Random and Uniform in all
dataset--model cells, with gains of up to $19.115\pp$ and
$3.280\pp$, respectively, while signal importance varies substantially
across cells.

    \item We propose \rcw\ and \csg\ to automatically allocate signal weights
from dataset--model reliability and candidate-pool context, respectively.
Both outperform Fixed in all nine cells, yielding positive adaptive
comparisons and gains of up to $6.621\pp$.

    \item Under matched mutation and query budgets, \rcw\ and \csg\ achieve
$16.17\%$ and $16.06\%$ failure discovery versus $6.06\%$ for the strongest
adapted neural-fuzzing baseline, and retain 11,508/12,058 similarity-based
failure clusters versus 6,210 for TensorFuzz.

    \item Wall-clock analysis further shows that
\rcw\ and \csg\ achieve substantially higher sustained failure throughput
than both Random and conventional neural-fuzzing baselines without
requiring greater online runtime; especially, TensorFuzz uses
$1.11\times$ the runtime of \rcw\ while achieving only $0.45\times$ its
failure throughput.

\end{itemize}

\section{Related Work}
\label{sec:related}

\noindent\textbf{Neural-symbolic learning.}
Neural-symbolic systems expose heterogeneous interfaces: DeepProbLog couples
neural predicates with probabilistic logic \citep{manhaeve2018deepproblog};
ISED supports arbitrary black-box programs \citep{solkobreslin2024ised};
DOLPHIN composes distributions and provenance \citep{naik2025dolphin}; NASR
uses symbolic correction \citep{cornelio2023nasr}; and NeSyCoCo executes
differentiable concept programs \citep{kamali2024nesycoco}. Their intermediate
states range from probabilities and sampled worlds to correction masks and
grounding tensors. We therefore standardize extracted risk semantics rather
than raw internal representations.

\noindent\textbf{Testing learned systems.}
DeepXplore, DeepTest, DeepHunter, TensorFuzz, and DLFuzz guide neural testing
using neuron coverage, metamorphic transformations, activation-space novelty,
or activation drift
\citep{pei2017deepxplore,tian2018deeptest,xie2019deephunter,odena2019tensorfuzz, guo2018dlfuzz}.
More recent approaches such as NSGen and CriticalFuzz further exploit
semantic or critical-neuron feedback \citep{huang2024neuron, bai2024criticalfuzz}. These methods provide effective guidance
for conventional neural networks, but their feedback is tied to neural
representations and does not directly expose grounding uncertainty, symbolic
constraint violations, or neural--symbolic execution boundaries. We therefore
compare against matched-feedback adaptations of DeepHunter, TensorFuzz,
DLFuzz, NSGen, and CriticalFuzz under identical mutations, semantic validity
constraints, frontier width, hardware, and query budgets. This isolates the
effect of search feedback while preserving the same NSAI testing environment.

\noindent\textbf{Adaptive multi-signal guidance.}
Multi-objective fuzzing can combine coverage, uncertainty, and distance, but
their relative importance is domain dependent. Our rank fusion removes scale
mismatch; \rcw\ learns dataset--model-level weights from calibrated reliability,
while \csg\ adapts weights to candidate-pool context during search. All policies
are evaluated on-policy because different frontier selections induce different
future candidate pools.

\section{Why a Unified Framework?}

\noindent\textbf{Interface mismatch.}
Let an NSAI system consist of a neural perceptual map $M_\theta$, a structured
state constructor $G$, and an executor $P$:
\begin{equation}
    F(x,q)=P\!\left(G(M_\theta(x),q)\right).
    \label{eq:pipeline}
\end{equation}
The optional query $q$ may be a relational question, arithmetic program, or
language command. Equation~\eqref{eq:pipeline} covers heterogeneous
intermediate states that are not directly interchangeable, so a coverage
metric designed for one interface cannot be applied uniformly across
Sort-of-CLEVR, Visual Sudoku, HWF-7, and ReaSCAN.

We therefore standardize \emph{risk semantics} rather than raw internal states.
Each adapter exposes available estimates of boundary pressure, grounding
uncertainty, representation drift, constraint violation, and novelty; missing
signals are masked, making one search procedure across heterogeneous
executors.

\noindent\textbf{Semantics before search.}
For a solved seed $(x,q,y)$, a task-specific mutation operator
$T$ produces $x'=T(x;\xi)$, $\xi$ denotes the mutation parameters, such
as perturbation magnitude, transformation type, or sampled replacement
instance. A task relation $\equiv_y$ certifies that the expected answer is
unchanged:
\begin{equation}
    x'\equiv_y x
    \quad\Longrightarrow\quad
    y(x',q)=y(x,q)=y.
\end{equation}
The oracle failure is
\begin{equation}
    f(x')=\ind[F(x',q)\neq y]\quad\text{subject to }x'\equiv_y x.
    \label{eq:failure}
\end{equation}
Photometric transformations are restricted by an RMS-$L_2$ ball, while
structured transformations preserve equivalence by construction. Invalid
candidates are discarded, so guidance affects search efficiency rather than
test validity.

\noindent\textbf{Threat model and objective.}
We assume grey-box forward access to output scores and selected intermediate
representations; gradients, parameter updates, and training data are not
required. Given query budget $Q$, policy $\pi$ maximizes failure yield
\begin{equation}
    J(\pi)=\mathbb{E}\!\left[
    \frac{1}{Q}\sum_{t=1}^{Q}f(x_t^\pi)
    \right].
    \label{eq:objective}
\end{equation}
All policies share seeds, candidate generators, validity filters, beam size,
and query caps, differing only in frontier selection. Thus, differences in
$J(\pi)$ reflect guidance.

\section{Method}

\subsection{Architecture-Neutral Guidance Signals}

For each valid candidate $x'$, the model-state adapter returns five available
signals $s(x')=[b,g,r,c,n]$. Let $z_k(x')$ be the score of answer $k$ and
$h(x')$ an adapter-selected intermediate representation.
\noindent\textbf{Boundary pressure}
measures proximity to an incorrect decision,
\begin{equation}
b(x')=\max_{k\neq y}z_k(x')-z_y(x'),
\end{equation}
with the corresponding execution margin used for structured outputs.
\textbf{Grounding uncertainty} $g$ measures instability at the
neural--symbolic interface using normalized entropy, prototype distance, or
execution uncertainty, depending on the exposed state.
\textbf{Representation drift} measures latent change,
\begin{equation}
r(x')=1-\frac{h(x)^\top h(x')}
{\|h(x)\|_2\|h(x')\|_2+\epsilon}.
\end{equation}
\textbf{Constraint violation} $c$ measures normalized symbolic inconsistency;
it is masked when no meaningful checker is available.
Finally, \textbf{Archive novelty} encourages unexplored states,
\begin{equation}
n(x')=\min_{a\in\mathcal A}
\left(1-\cos(h(x'),a)\right).
\end{equation}
Although their implementations depend on the adapter, all five signals share
the same risk interpretation: larger values receive higher search priority.

\subsection{Pool-Wise Rank Fusion}

Because the five raw signals have heterogeneous scales, we rank candidates
within the current valid pool $C_t=\{x'_1,\ldots,x'_m\}$. Let $s_{ij}$ be
signal $j\in\{1,\ldots,5\}$ for candidate $x'_i$. Its percentile rank is
\begin{equation}
    \rho_{ij}=
    \frac{1}{m-1}\sum_{\ell=1}^{m}\ind[s_{\ell j}<s_{ij}],
    \qquad \rho_{ij}\in[0,1],
    \label{eq:rank}
\end{equation}
where larger $\rho_{ij}$ indicates higher rank; ties are resolved
deterministically.

To handle unavailable signals, let
$a=(a_1,\ldots,a_5)\in\{0,1\}^5$, with $a_j=1$ if signal $j$ is available,
and let $w\in\Delta^4$, where
$\Delta^4=\{w\in\mathbb{R}_{\geq0}^5:\sum_jw_j=1\}$.
The masked and normalized fusion score is
\begin{equation}
    S_w(x_i')=\sum_{j=1}^{5}\bar w_j\rho_{ij},\qquad
    \bar w_j=
    \begin{cases}
    \dfrac{a_jw_j}{\sum_{k=1}^{5}a_kw_k},
        & \sum_{k=1}^{5}a_kw_k>0,\\[4pt]
    0, & \text{otherwise}.
    \end{cases}
    \label{eq:fusion}
\end{equation}
Larger $S_w$ gives higher frontier priority. If all nominal weight is assigned
to unavailable signals, all candidates tie and a seeded stable tie-break is
used.

Fixed uses $w^0=(.35,.20,.20,.15,.10)$ for boundary, grounding, drift,
constraint, and novelty. Uniform assigns equal weight to available signals;
one-hot policies retain one signal; leave-one-out policies remove one signal
and renormalize the remaining Fixed weights; Random ignores $S_w$ and randomly
orders the frontier.

\begin{proposition}[Monotone-scale invariance]
For any strictly increasing transformations $\phi_j$, replacing every raw
signal $s_{ij}$ by $\phi_j(s_{ij})$ leaves all percentile ranks
$\rho_{ij}$ unchanged and therefore preserves the fused scores and selected
frontier.
\end{proposition}

\begin{proposition}[Availability safety]
For any two nominal weight vectors whose available coordinates are
proportional, masked normalization produces the same effective weights
$\bar w$ and hence the same candidate ordering. Unavailable signals receive
zero effective weight and cannot affect selection.
\end{proposition}

Strictly monotone transformations preserve within-pool ordering, while masking
preserves relative weight among available signals; proofs are given in
Appendix~\ref{app:proofs}.

\subsection{Reliability-Calibrated Weighting}

\rcw\ learns one signal-weight vector per dataset--model cell from calibration
candidates $\{x'_i,f_i\}_{i=1}^{N_{\mathrm{cal}}}$, where
$f_i\in\{0,1\}$ denotes failure. For signal $j$, we measure failure-ranking
ability by
\begin{equation}
    A_j=\AUC(\{s_{ij}\}_{i=1}^{N_{\mathrm{cal}}},
             \{f_i\}_{i=1}^{N_{\mathrm{cal}}}),
\end{equation}
and define reliability utility
\begin{equation}
    u_j=\max(2A_j-1,0),
    \label{eq:gini}
\end{equation}
so $u_j\in[0,1]$ and signals no better than random receive zero positive
evidence.

Let
\[
\Delta_a=
\left\{
w\in\mathbb{R}_{\geq0}^{5}:
\sum_{j=1}^{5}w_j=1,\;
w_j=0\ \text{if}\ a_j=0
\right\},
\]
and let $w_a^0$ be the availability-masked Fixed prior. \rcw\ solves
\begin{equation}
 w^{\mathrm{RCW}}
 =\arg\max_{w\in\Delta_a}
 \left\{
 \langle w,u\rangle
 -\frac{1}{\beta}D_{\mathrm{KL}}(w\|w_a^0)
 \right\},
 \label{eq:rcw-opt}
\end{equation}
where $\langle w,u\rangle=\sum_jw_ju_j$ rewards reliable signals,
$D_{\mathrm{KL}}$ regularizes deviation from the prior, and $\beta>0$
controls reweighting strength.

The closed-form solution is
\begin{equation}
    w_j^{\mathrm{RCW}}=
    \frac{a_jw_j^0\exp(\beta u_j)}
    {\sum_{k=1}^{5}a_kw_k^0\exp(\beta u_k)},
    \label{eq:rcw}
\end{equation}
with $\beta=3$. Thus, reliable signals receive more weight while the prior
limits extreme allocations; $w^{\mathrm{RCW}}$ remains fixed within each
dataset--model cell.

\begin{proposition}[Optimality and stability of \rcw]
If $w_{a,j}^0>0$ for every available signal, Equation~\eqref{eq:rcw-opt} has
the unique maximizer in Equation~\eqref{eq:rcw}. For utility vectors $u,u'$,
any available signals $j,k$ satisfy
\begin{equation}
 \left|
 \log\frac{w_j(u)/w_k(u)}
          {w_j(u')/w_k(u')}
 \right|
 \leq 2\beta\|u-u'\|_\infty .
 \label{eq:rcw-stability}
\end{equation}
\end{proposition}

Hence, \rcw\ is uniquely defined and its relative weights are stable to
calibration perturbations; derivations appear in Appendix~\ref{app:proofs}.

\subsection{Contextual Signal Gating}

Because cell-level weights cannot adapt to changing candidate pools, \csg\
builds a round-level context vector $v_t$ from ranked-signal statistics,
pairwise correlations, signal availability, search progress, and task/model
descriptors. A linear gate predicts
\begin{equation}
    w_t^{\mathrm{CSG}}
    =\softmax_a(Wv_t+\eta),
    \label{eq:csg}
\end{equation}
where $W$ is learnable, $\eta=\log w^0$, and $\softmax_a$ masks unavailable
signals. Thus, $w_t^{\mathrm{CSG}}\in\Delta^4$ may vary across rounds.

For a sampled failure/non-failure pair $(p,n)$ with $f_p=1$ and $f_n=0$,
let $\rho_p,\rho_n\in[0,1]^5$ be their ranked signal vectors and define
\begin{equation}
    m_{pn}=(w_t^{\mathrm{CSG}})^\top(\rho_p-\rho_n),
\end{equation}
so $m_{pn}>0$ ranks the failure higher.

The gate minimizes
\begin{align}
    \mathcal{L}
    &=\frac{1}{|\mathcal{P}|}\sum_{(p,n)\in\mathcal{P}}
      \log(1+\exp(-m_{pn})) \nonumber\\
    &\quad+\lambda\|W\|_F^2+
    \gamma D_{\mathrm{KL}}(w_a^0\|w_t^{\mathrm{CSG}}),
    \label{eq:loss}
\end{align}
where $\mathcal{P}$ is the sampled pair set, $\lambda$ controls $\ell_2$
regularization, and $\gamma$ regularizes toward the masked prior. We sample at
most 64 pairs per pool and train for 150 epochs with learning rate $.03$,
$\lambda=10^{-4}$, and $\gamma=.05$.

\begin{proposition}[Misranking surrogate]
For every failure/non-failure pair,
\begin{equation}
    \ind[m_{pn}\leq0]\leq
    \frac{\log(1+\exp(-m_{pn}))}{\log 2}.
\end{equation}
\end{proposition}

Hence, the pairwise loss upper-bounds empirical misranking. This is a local
ranking guarantee, so all policies are evaluated on-policy.

\subsection{Guided Search}

We instantiate Equation~\eqref{eq:objective} with bounded beam search. At round
$t$, each parent $u\in\mathcal{B}_t$ is mutated by $T(u;\xi)$, and semantic
filtering yields
\begin{equation}
    C_t=\{T(u;\xi):u\in\mathcal{B}_t\}
    \cap\{x':x'\equiv_y x\},
    \label{eq:candidate-pool}
\end{equation}
where each retained candidate is executed once and counts toward the query
budget.

With beam width $B$ and signal weights $w_t$, score-based policies retain
\begin{equation}
    \mathcal{B}_{t+1}=
    \operatorname*{arg\,topB}_{x_i'\in C_t}S_{w_t}(x_i'),
    \label{eq:beam-update}
\end{equation}
while Random selects $B$ candidates by seeded permutation. All policies share
seeds, mutations, validity checks, beam width, and query cap, differing only in
parent selection.

Evaluation is \emph{on-policy}: different parents induce different future
candidate pools. Random-frontier pools are used only for \rcw/\csg\
calibration; reported results come from fresh policy trajectories. The archive
$\mathcal{A}_t$ stores the seed and selected parents, and failures do not
terminate search. For $m_t=|C_t|$ and representation dimension $d$, rank fusion
costs $O(5m_t\log m_t)$ and exact novelty
$O(m_t|\mathcal{A}_t|d)$. Additional details appear in
Appendix~\ref{app:proofs}.

\section{Experimental Setup}

\noindent\textbf{Evaluation configuration.}
We evaluate ISED and DOLPHIN on Sort-of-CLEVR, Visual Sudoku, and HWF-7; NASR
on Visual Sudoku and HWF-7; and NeSyCoCo on ReaSCAN, yielding nine compatible
dataset--model cells. The tasks cover relational VQA
\citep{santoro2017relation}, $9\times9$ Sudoku \citep{wang2019satnet},
handwritten arithmetic, and compositional visual grounding
\citep{wu2021reascan}. Each adapter defines its mutation space, semantic oracle,
and available signals; only initially correct inputs are fuzzed.
Each cell evaluates 20 policies: Random, Uniform, Fixed
$(.35,.20,.20,.15,.10)$, five one-hot, five leave-one-out, \rcw, \csg,
and five matched neural-fuzzing baselines---DeepHunter, TensorFuzz, DLFuzz,
NSGen-transfer, and CriticalFuzz. All policies use the same 20 initially
correct inputs per cell, mutation operators, semantic validity checks, model
checkpoints, beam width eight, hardware, and 256 candidate evaluations per
input, differing only in search feedback and frontier selection. Ten paired
attack seeds therefore yield
$9\times20\times10=1{,}800$ on-policy runs, corresponding to
$9{,}216{,}000$ candidate evaluations. Adapter and implementation details are
given in Appendices~\ref{app:adapters} and~\ref{app:repro}.

\noindent\textbf{Calibration and metrics.}
Random-frontier traces are grouped into candidate pools, with alternating pools
used for calibration and screening and at most 2,000 retained per cell.
\rcw\ estimates single-signal AUROC, while \csg\ is fit on calibration pools;
all policies are then rerun from scratch on-policy. The primary metric is
candidate-level oracle failure discovery rate,
\begin{equation}
\widehat J=\frac{\sum_i f(x_i')}{N_{\mathrm{valid}}}.
\end{equation}
We report means over the same ten paired attack seeds for all policies.
We additionally report exact-output deduplication, activation-similarity
failure clusters, and wall-clock runtime, TTF, $Q_{\mathrm{first}}$, and
failure throughput.
Experiments use three NVIDIA RTX A6000 48 GB GPUs, a 64-thread CPU
environment, PyTorch, and CUDA.

\begin{figure*}[t]
    \centering
    \includegraphics[width=0.9\linewidth]{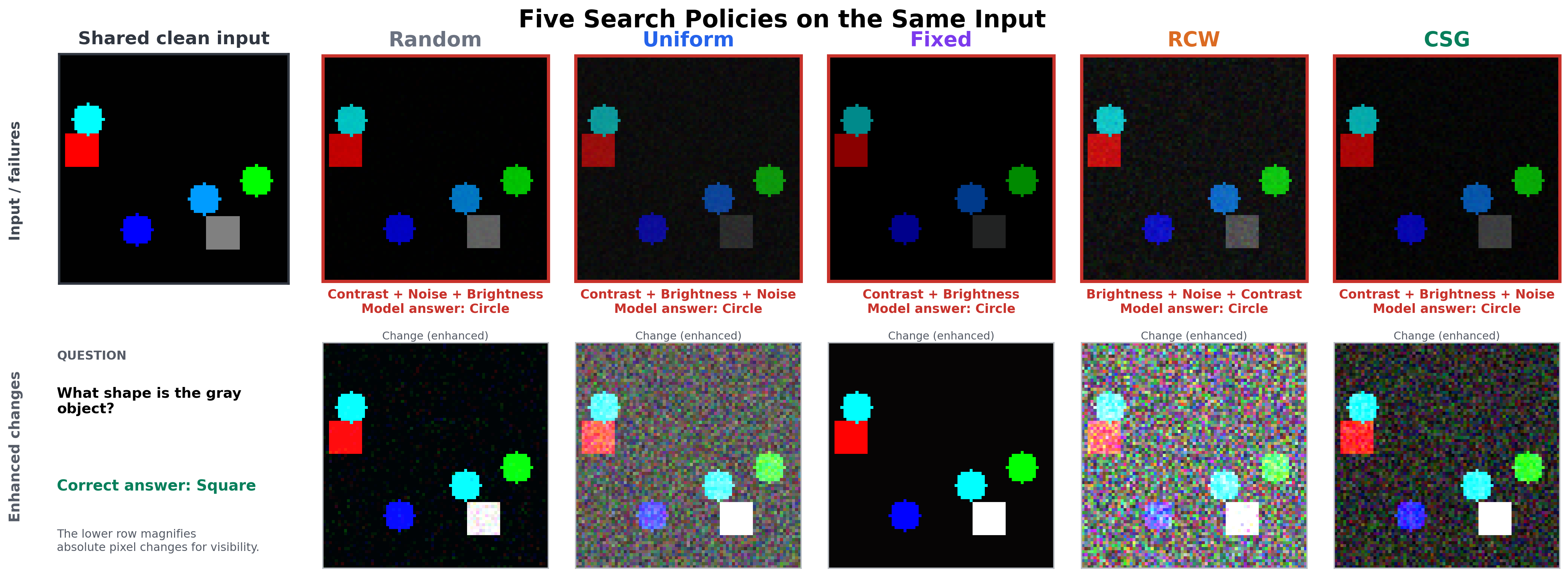}
    \caption{Five search policies on the Sort-of-CLEVR. All mutations
preserve the oracle answer \emph{Square} but induce the prediction
\emph{Circle}; the bottom row visualizes enhanced pixel differences.}
    \label{fig:search-example}
\end{figure*}

\section{Fuzzing Results}
\subsection{Qualitative Search Example}

Figure~\ref{fig:search-example} shows how different policies explore the same
initially correct Sort-of-CLEVR seed. All mutations preserve the oracle answer
\emph{Square} but induce the incorrect prediction \emph{Circle}, illustrating
that frontier selection reaches different regions of the valid mutation space.
The lower row shows enhanced absolute pixel differences from the clean input.

\begin{table*}[t]
\centering
\scriptsize
\caption{Mean failure discovery rate (\%) across ten paired attack seeds.}
\label{tab:main}
\resizebox{\textwidth}{!}{
\begin{tabular}{ll|rrrrr|rrr|rr}
\toprule
Dataset & Model &
DeepHunter & TensorFuzz & DLFuzz & NSGen-T & CriticalFuzz &
Random & Uniform & Fixed &
\rcw & \csg \\
\midrule

Sort-of-CLEVR & ISED
& 0.18 & 1.91 & 1.31 & 0.39 & 0.14
& 1.127 & 11.693 & 12.675
& \textbf{13.832} & 13.526 \\

Sort-of-CLEVR & DOLPHIN
& 1.35 & 1.60 & 1.50 & 0.59 & 1.33
& 2.108 & 13.723 & 14.335
& 14.914 & \textbf{15.557} \\

Visual Sudoku & ISED
& 4.69 & 5.72 & 5.14 & 3.95 & 4.49
& 6.645 & 12.791 & 14.133
& 16.062 & \textbf{16.629} \\

Visual Sudoku & DOLPHIN
& 0.00 & 0.00 & 0.00 & 0.00 & 0.00
& 1.395 & 4.002 & 4.908
& \textbf{5.764} & 5.740 \\

Visual Sudoku & NASR
& 9.96 & 9.08 & 10.31 & 10.02 & 9.59
& 9.598 & 18.023 & 21.303
& 23.598 & \textbf{23.740} \\

HWF-7 & ISED
& 3.30 & 7.23 & 8.12 & 0.80 & 2.93
& 0.619 & 5.904 & 6.725
& 9.680 & \textbf{9.934} \\

HWF-7 & DOLPHIN
& 3.67 & 9.30 & 10.20 & 12.32 & 3.67
& 2.582 & 15.232 & 18.098
& \textbf{24.719} & 23.123 \\

HWF-7 & NASR
& 4.77 & 14.88 & 10.00 & 18.09 & 5.96
& 3.270 & 21.447 & 22.385
& \textbf{26.650} & 25.633 \\

ReaSCAN & NeSyCoCo
& 0.74 & 4.84 & 3.89 & 2.25 & 5.57
& 1.074 & 9.889 & 10.211
& 10.283 & \textbf{10.664} \\

\bottomrule
\end{tabular}
}
\end{table*}

\subsection{Structured Guidance Within \method}

Table~\ref{tab:main} first compares the internal \method\ policies. Fixed guidance
outperforms Random in all nine dataset--model cells, often by a large margin: 12.68\% versus 1.13\% on Sort-of-CLEVR/ISED, 18.10\% versus 2.58\%
on HWF-7/DOLPHIN, and 10.21\% versus 1.07\% on ReaSCAN/NeSyCoCo. Since the two
policies share the same seeds, mutation operators, validity checks, beam width,
and query budgets, while Random selects the frontier without using the five
guidance signals, these gains show that boundary pressure, grounding
uncertainty, representation drift, constraint violation, and novelty provide
effective information for steering the search toward failure-prone regions.
Thus, structured five-signal guidance is substantially more effective than
unguided random frontier selection.

Fixed also outperforms Uniform in every cell. Because all five signals are
rank-normalized before fusion, this gap reflects unequal signal usefulness
rather than numerical scale. Moreover, both \rcw\ and \csg\ further improve
over Fixed in every cell, with gains of up to $6.621\pp$. These comparisons
establish the effectiveness of structured and adaptive guidance within the
common \method\ search framework. We next compare against established
neural-fuzzing feedback under the same evaluation configuration.

\subsection{Comparison with Neural-Fuzzing Baselines}
\label{sec:nn-baselines}

Table~\ref{tab:main} compares \rcw/\csg\ with five established
neural-fuzzing feedback policies under the same inputs, mutations, semantic
constraints, beam width, checkpoints, hardware, and query budgets. Conventional
neural feedback remains useful, but is consistently weaker than guidance
designed around the neural--symbolic interface. Across all nine cells,
TensorFuzz is the strongest neural baseline in aggregate, with a mean
candidate-level failure discovery rate of 6.06\%, whereas \rcw\ and \csg
reach 16.17\% and 16.06\%, corresponding to $2.67\times$ and $2.65\times$
higher failure discovery.

The advantage is also broadly distributed across individual interfaces.
On Visual Sudoku/NASR, the strongest neural baseline reaches 10.31\%, compared
with 23.598\% and 23.740\% for \rcw/\csg; on HWF-7/DOLPHIN,
NSGen-transfer reaches 12.32\%, compared with 24.719\% and 23.123\%; and on
ReaSCAN/NeSyCoCo, CriticalFuzz reaches 5.57\%, compared with 10.283\% and
10.664\% for \rcw/\csg. Visual Sudoku/DOLPHIN is particularly
sparse: \rcw\ discovers failures while all five neural-fuzzing baselines find
none. These results indicate that activation coverage, distance, and
neuron/path novelty provide useful exploration signals, but do not fully
capture risks arising at grounding, constraint, and neural--symbolic execution
interfaces.

\begin{table}[t]
\centering
\scriptsize
\caption{Failure diversity across all nine dataset--model cells and ten paired
attack seeds.}
\label{tab:failure-diversity}
\begin{tabular}{lrrr}
\toprule
Policy & Raw failures & Exact unique & Similarity clusters \\
\midrule
DeepHunter     & 14,670 & 5,300 & 5,450 \\
TensorFuzz     & 27,940 & 5,610 & 6,210 \\
DLFuzz         & 25,840 & 5,390 & 5,850 \\
NSGen-transfer & 24,780 & 5,330 & 5,740 \\
CriticalFuzz   & 17,240 & 5,370 & 5,620 \\
\midrule
\rcw            & 74,506 & 10,719 & 11,508 \\
\csg            & 74,003 & \textbf{11,311} & \textbf{12,058} \\
\bottomrule
\end{tabular}
\end{table}

\noindent\textbf{Failure diversity after deduplication.}
A higher candidate-level failure rate could result from repeatedly discovering
the same or highly similar failures. Table~\ref{tab:failure-diversity} therefore reports two complementary
diversity summaries. Exact deduplication retains one failure for each
clean-seed/erroneous-output pair, while activation-similarity clustering
partitions failures with the same clean seed and erroneous output according
to their activation embeddings.

The advantage of \rcw/\csg\ remains after both forms of deduplication.
Among the neural-fuzzing baselines,
TensorFuzz retains the largest number of exact-unique failures (5,610) and
similarity clusters (6,210). In comparison, \rcw\ retains 10,719 exact-unique
failures and 11,508 similarity clusters, while \csg\ retains 11,311 and
12,058, respectively. Relative to TensorFuzz, \rcw/\csg\ therefore discover
$1.91\times/2.02\times$ as many exact-unique failures and
$1.85\times/1.94\times$ as many similarity-clustered failure behaviors.
These results indicate that the higher failure-discovery rate of adaptive
NSAI guidance is not explained simply by repeatedly generating near-duplicate
failures.

The diversity advantage is also robust to the clustering threshold: for
cosine thresholds 0.95, 0.98, and 0.99, the strongest neural baseline retains
5,910, 6,210, and 6,590 clusters, whereas \rcw\ retains 10,982, 11,508,
and 12,021 and \csg\ retains 11,659, 12,058, and 12,638, respectively. The complete threshold
sensitivity results are provided in Appendix~\ref{app:dedup-sensitivity}. These results show that the higher failure-discovery rate of adaptive NSAI guidance is not explained simply by repeatedly generating near-duplicate failures. We interpret these clusters as distinct observed failure behaviors rather than causally verified independent root causes.

\begin{table}[h]
\centering
\scriptsize
\caption{Single-signal failure discovery rates (\%). }
\label{tab:single-full}
\begin{tabular}{l|rrrrr}
\toprule
Dataset--model & B & G & R & C & N \\
\midrule
Sort-of-CLEVR--ISED & \textbf{12.564} & 11.947 & 5.801 & 1.885 & 5.916 \\
Sort-of-CLEVR--DOLPHIN & 13.193 & \textbf{13.327} & 12.315 & 1.424 & 3.817 \\
Visual Sudoku--ISED & \textbf{15.815} & 5.387 & 8.053 & 10.100 & 7.680 \\
Visual Sudoku--DOLPHIN & \textbf{5.676} & 2.141 & 1.209 & 1.615 & .895 \\
Visual Sudoku--NASR & \textbf{23.000} & 8.473 & 11.844 & 15.094 & 9.781 \\
HWF-7--ISED & 8.742 & 5.824 & \textbf{8.954} & 2.979 & 4.934 \\
HWF-7--DOLPHIN & \textbf{22.355} & 11.441 & 19.125 & 3.129 & 10.533 \\
HWF-7--NASR & \textbf{24.820} & 11.791 & 22.848 & 3.549 & 11.611 \\
ReaSCAN--NeSyCoCo & \textbf{9.992} & 0.787 & 9.689 & 0.693 & 4.170 \\
\bottomrule
\end{tabular}
\end{table}

\subsection{Signal Importance Is Heterogeneous}
\label{sec:ablation-main}

Having established that NSAI-specific guidance outperforms both internal
baselines and conventional neural-fuzzing feedback, we next examine why a
single fixed signal allocation is insufficient. Single-signal and leave-one-out (LOO) diagnostics reveal that the five guidance
signals contribute very differently across dataset--model cells. A one-hot
policy measures the stand-alone search strength of one signal, while a LOO
policy removes one signal from Fixed and renormalizes the remaining mass,
revealing whether that signal is helpful under the fixed combination.

\begin{table*}[t]
\centering
\scriptsize
\caption{Leave-one-out (LOO) signal ablation (\%). }
\label{tab:loo-main}
\begin{tabular}{l|rrrrr}
\toprule
Dataset--model & $-$B & $-$G & $-$R & $-$C & $-$N \\
\midrule
Sort-of-CLEVR--ISED & 10.632 & 12.305 & 12.577 & \textbf{12.760} & 12.659 \\
Sort-of-CLEVR--DOLPHIN & 12.668 & 13.995 & 14.117 & 14.305 & \textbf{14.630} \\
Visual Sudoku--ISED & 9.627 & 14.863 & 14.736 & \textbf{15.031} & 14.055 \\
Visual Sudoku--DOLPHIN & 1.727 & 4.639 & \textbf{5.123} & 4.719 & 5.012 \\
Visual Sudoku--NASR & 13.500 & 20.764 & 21.059 & \textbf{21.881} & 21.059 \\
HWF-7--ISED & 5.514 & 7.416 & 5.934 & \textbf{7.711} & 6.484 \\
HWF-7--DOLPHIN & 13.986 & \textbf{18.998} & 16.873 & 17.977 & 18.496 \\
HWF-7--NASR & 17.266 & \textbf{24.994} & 20.809 & 21.240 & 21.783 \\
ReaSCAN--NeSyCoCo & 8.922 & \textbf{10.447} & 9.646 & 10.127 & 9.857 \\
\bottomrule
\end{tabular}
\end{table*}

Table~\ref{tab:single-full} shows that the stand-alone strength of the five
signals varies substantially across model--task cells. Boundary is the strongest
single signal in seven of nine cells, reaching 23.000\% on Visual Sudoku/NASR,
22.355\% on HWF-7/DOLPHIN, and 24.820\% on HWF-7/NASR. However, it is not
universally dominant: Grounding performs best on Sort-of-CLEVR/DOLPHIN
(13.327\%), while Drift is strongest on HWF-7/ISED (8.954\%). Other signals can
be much weaker in isolation, indicating that no single guidance signal is
uniformly effective across heterogeneous NSAI systems.

Table~\ref{tab:loo-main} further reveals that signal usefulness also changes
when signals are combined. Removing Constraint gives the best LOO result in
four cells, including Sort-of-CLEVR/ISED and Visual Sudoku/NASR, whereas
removing Grounding is best for HWF-7/DOLPHIN, HWF-7/NASR, and
ReaSCAN/NeSyCoCo. Sort-of-CLEVR/DOLPHIN instead benefits most from removing
Novelty, while Visual Sudoku/DOLPHIN favors removing Drift. These results show
that a signal that is strong individually is not necessarily equally useful in
a multi-signal combination. More importantly, both Single and LOO analyses are
post-hoc diagnostics and do not determine the appropriate signal weights in
advance, motivating \rcw\ and \csg\ to learn signal allocations from
dataset--model reliability and candidate-pool context, respectively.

\subsection{Adaptive Weighting Learns Effective Signal Allocations}

The preceding diagnostics show that signal importance varies substantially
across model--task cells, making weight allocation itself the key challenge.
\rcw\ learns a cell-level allocation from calibrated signal reliability,
whereas \csg\ further adapts the allocation to the current candidate-pool
context.

Table~\ref{tab:main} shows that both \rcw\ and \csg\ outperform Fixed in all
nine dataset--model cells, yielding positive gains in all 18 adaptive
comparisons. The largest improvements occur on HWF-7: for DOLPHIN, \rcw\
raises failure discovery from 18.098\% to 24.719\% ($+6.621\pp$) and \csg\
to 23.123\% ($+5.025\pp$); for NASR, the corresponding gains are
$+4.265\pp$ and $+3.248\pp$. On HWF-7/ISED, \rcw\ and \csg\ improve Fixed
by $2.955\pp$ and $3.209\pp$, respectively.
The gains also extend across other reasoning interfaces. On Visual
Sudoku/NASR, \rcw\ and \csg\ improve Fixed by $2.295\pp$ and $2.437\pp$;
on Sort-of-CLEVR/DOLPHIN, \csg\ gains $1.222\pp$; and on ReaSCAN, \csg\ still improves Fixed from 10.211\% to
10.664\%. These results show that adaptive allocation remains effective across
different tasks, models, and signal-availability patterns.

Neither mechanism is uniformly superior: \rcw\ achieves the higher rate in
four cells and \csg\ in five. This reflects their complementary adaptation
granularities---\rcw\ captures dataset--model-level reliability, whereas
\csg\ responds to candidate-pool context during search. Complete adaptive
gains, learned \rcw\ allocations, and cross-dataset analysis are provided in
Appendix~\ref{app:full-results}.

\subsection{Wall-Clock Fuzzing Efficiency}
\label{sec:runtime-efficiency}

Table~\ref{tab:runtime} compares online search time, capped wall-clock time
to first failure (TTF), capped first-failure query count
$Q_{\mathrm{first}}$, and failure-discovery throughput. To make the internal
and neural-fuzzer comparisons directly readable in one table, each metric is
normalized to \rcw\ within the same dataset--model cell and then geometrically
averaged across cells. Lower values are better for Runtime, TTF, and
$Q_{\mathrm{first}}$, while higher values are better for Fail./s. The complete
timing protocol is given in Appendix~\ref{app:runtime}.

\begin{table}[t]
\centering
\scriptsize
\caption{Wall-clock efficiency of matched neural-fuzzing baselines and
\method\ policies.}
\label{tab:runtime}
\begin{tabular}{l|rrrr}
\toprule
Policy &
Runtime ($\times$)$\downarrow$ &
TTF ($\times$)$\downarrow$ &
$Q_{\mathrm{first}}$ ($\times$)$\downarrow$ &
Fail./s ($\times$)$\uparrow$ \\

\midrule
Random          & 0.71 & 0.74 & 1.08 & 0.20 \\
Uniform         & 1.02 & 1.04 & 1.00 & 0.81 \\
Fixed           & 1.01 & 1.01 & 1.01 & 0.91 \\
\midrule
DeepHunter      & 1.09 & 1.15 & 1.09 & 0.17 \\
TensorFuzz      & 1.11 & 1.15 & 1.06 & 0.45 \\
DLFuzz          & 1.11 & 1.16 & 1.06 & 0.40 \\
NSGen-transfer  & 1.10 & 1.09 & 1.02 & 0.23 \\
CriticalFuzz    & 1.09 & 1.11 & 1.02 & 0.22 \\
\midrule
\rcw            & \textbf{1.00} & \textbf{1.00} &
                   \textbf{1.00} & \textbf{1.00} \\
\csg            & 1.01 & 1.02 & 1.02 & 0.98 \\
\bottomrule
\end{tabular}
\end{table}

Random has the lowest online runtime ($0.71\times$ \rcw) because it performs
no guidance computation, but its failure throughput is only $0.20\times$
that of \rcw. Uniform and Fixed incur nearly the same runtime as the adaptive
methods, yet achieve only $0.81\times$ and $0.91\times$ \rcw's failure
throughput. Thus, most of the guided-search cost arises from common signal
extraction and ranking rather than adaptive weighting itself.

The neural-fuzzing baselines do not provide a wall-clock advantage over the
adaptive methods. TensorFuzz, the strongest neural baseline in aggregate
failure discovery, requires $1.11\times$ the runtime and $1.15\times$ the
capped TTF of \rcw\ while achieving only $0.45\times$ its failure throughput.
The remaining neural baselines exhibit similarly lower throughput
($0.17$--$0.40\times$) despite comparable or higher online runtime.
\csg\ remains close to \rcw\ in both cost and effectiveness, requiring only
$1.01\times$ its runtime and achieving $0.98\times$ its throughput.
Overall, adaptive NSAI-specific guidance improves sustained failure discovery
without relying on greater online wall-clock expenditure.

\section{Conclusion}

We presented \method, a semantics-aware fuzzing framework that maps heterogeneous NSAI states into five availability-aware, rank-normalized risk signals. Across nine dataset--model cells, structured guidance consistently outperforms Random and Uniform, while matched comparisons further show that \rcw\ and \csg\ substantially outperform established neural-fuzzing feedback, achieving 16.17\% and 16.06\% aggregate failure discovery versus 6.06\% for the strongest neural baseline and retaining substantially more exact-unique and similarity-clustered failure behaviors after deduplication. Single-signal and leave-one-out analyses reveal substantial heterogeneity in signal importance across tasks and models, motivating adaptive signal allocation; accordingly, \rcw\ and \csg\ adapt to dataset--model reliability and candidate-pool context, respectively, and improve over Fixed across all evaluated cells. Wall-clock results further show that these gains do not rely on greater online computation: TensorFuzz requires $1.11\times$ the runtime of \rcw\ while achieving only $0.45\times$ its failure throughput. Overall, these results demonstrate that a common fuzzing interface is practical across heterogeneous NSAI systems and that effective and efficient testing benefits from neural--symbolic-specific risk guidance with adaptive signal allocation.

\label{sec:main-end}

\section*{Reproducibility Statement}
The experimental unit, compatibility matrix, policy definitions, mutation
oracles, query budget, beam width, ten paired attack seeds, and calibration
protocol are specified in Sections~5--6 and Appendix~\ref{app:repro}.
Appendix~\ref{app:adapters} defines the adapter-level signals and semantic
validity conditions. Tables~\ref{tab:single-full} and~\ref{tab:loo-main}
report the complete single-signal and leave-one-out results, while
Appendix~\ref{app:full-results} provides the complete adaptive improvements
and learned \rcw\ allocations. Appendix~\ref{app:dedup-sensitivity} reports
failure-deduplication threshold sensitivity, and
Appendix~\ref{app:runtime} specifies the wall-clock measurement protocol. Appendix~\ref{app:proofs} contains supplementary derivations and method
details. 
The anonymous artifact includes the adapters, configurations, attack seeds,
candidate-level summaries, timing logs, and analysis code required to
reproduce the reported results and tables.

\section*{Ethics Statement}
This work studies defensive reliability testing on synthetic or public
benchmarks and involves no human subjects or private data. Guided fuzzing can
make failures easier to find; semantic mutation constraints are reported so
that discovered cases remain valid tests rather than unrestricted adversarial
examples. Rates must not be used as cross-dataset robustness rankings because
candidate distributions differ by task.

\section*{AI Use Statement}
Generative AI tools were used to assist with language editing, document
organization, and LaTeX drafting. The authors remain responsible for the
scientific claims and verified the implementation descriptions, formulas,
experimental records, and numerical results against the retained artifacts.
AI tools were not treated as a source of experimental evidence.

\bibliography{references}
\bibliographystyle{iclr2027_conference}

\appendix
\section{Additional Derivations and Formal Details}
\label{app:proofs}

\subsection{Fuzzing as a constrained finite-horizon process}
Let $\sigma_t=(\mathcal{B}_t,\mathcal{A}_t,t,q_t)$ denote the search state,
where $q_t$ is the remaining query budget. A policy selects a beam from the
valid pool; the mutator and model jointly induce the next-state kernel
$P_T(\sigma_{t+1}\mid\sigma_t,\mathcal{B}_{t+1})$. The constrained return is
\begin{equation}
 V^\pi(\sigma_0)=\mathbb{E}_{\pi,P_T}\!\left[
 \sum_{t=0}^{H-1}\sum_{x\in C_t} f(x)
 \ \middle|\ C_t\subseteq\mathcal{V}(x_0),\;
 \sum_t|C_t|\leq Q\right],
 \label{eq:mdp-return}
\end{equation}
where $\mathcal{V}(x_0)=\{x:x\equiv_yx_0\}$ is the semantic validity set.
The score is not itself a reward model: its top-$B$ operator defines $\pi$.
Offline re-ranking fixes $C_t$ from a behavior policy and therefore does not
integrate over the transition kernel induced by a new frontier.

\subsection{RCW from a Gibbs posterior}
For available signals, introduce a Lagrange multiplier $\nu$ for
$\sum_jw_j=1$. Differentiating Equation~\eqref{eq:rcw-opt} gives
\begin{equation}
 u_j-\beta^{-1}\!\left(\log\frac{w_j}{w_{a,j}^0}+1\right)+\nu=0.
\end{equation}
Thus $w_j\propto w_{a,j}^0e^{\beta u_j}$. The normalizer is the partition
function $Z(u)=\sum_{k:a_k=1}w_{a,k}^0e^{\beta u_k}$, yielding
$w_j=\partial\log Z/(\beta\partial u_j)$. Its Jacobian is
\begin{equation}
 \frac{\partial w_j}{\partial u_k}
 =\beta w_j(\ind[j=k]-w_k).
 \label{eq:rcw-jacobian}
\end{equation}
Large $\beta$ reacts sharply to AUROC estimation error, while
$\beta\rightarrow0$ recovers the masked prior.

\subsection{Contextual ranking risk}
Let $\mathcal{D}$ be the calibration distribution over candidate pools and
$\mathcal{P}(C)$ the failure/non-failure pairs in pool $C$. The population
objective approximated by \csg\ is
\begin{equation}
 \mathcal{R}(W)=\mathbb{E}_{C\sim\mathcal{D}}
 \mathbb{E}_{(p,n)\sim\mathcal{P}(C)}
 \ell\!\left(w_W(v_C)^\top(\rho_p-\rho_n)\right)
 +\lambda\|W\|_F^2+\gamma
 D_{\mathrm{KL}}(w_a^0\|w_W(v_C)),
 \label{eq:population-risk}
\end{equation}
where $\ell(m)=\log(1+e^{-m})$. Since
$\ind[m\leq0]\leq\ell(m)/\log2$, logistic risk controls local pairwise
misranking. It does not imply global regret under
Equation~\eqref{eq:mdp-return}; the on-policy matrix evaluates long-horizon
behavior.

\section{Adapter and Oracle Specifications}
\label{app:adapters}

\begin{table}[h]
\centering
\small
\caption{Compatibility-driven evaluation matrix and semantic mutation oracle.}
\begin{tabular}{p{.18\textwidth}p{.28\textwidth}p{.42\textwidth}}
\toprule
Dataset & Models & Mutation/oracle \\
\midrule
Sort-of-CLEVR & ISED, DOLPHIN & Imaging changes; fixed generated scene,
question, and answer \\
Visual Sudoku & ISED, DOLPHIN, NASR & Digit rerendering and board symmetries;
fixed solution \\
HWF-7 & ISED, DOLPHIN, NASR & Same-class real glyph replacement and bounded
imaging; fixed expression \\
ReaSCAN & NeSyCoCo & Imaging changes; fixed scene graph, command, and referent \\
\bottomrule
\end{tabular}
\end{table}

\paragraph{Sort-of-CLEVR.}
Scenes contain colored geometric objects and binary relational questions.
Mutations alter brightness, contrast, color temperature, and bounded noise
without changing object count, identity, spatial relation, or question. ISED
and DOLPHIN expose answer distributions and relational states. Boundary is the
correct-answer execution margin; grounding is object or execution uncertainty;
drift uses relation states; constraint measures mass inconsistent with the
generated scene program.
Recorded clean QA accuracy is 100.0\% for ISED and 99.63\% for DOLPHIN.

\paragraph{Visual Sudoku.}
A grid is rendered from digit glyphs while the puzzle and unique solution
remain fixed. Exact row/column-preserving transformations and bounded glyph
changes define validity. ISED and DOLPHIN connect digit perception to a solver;
NASR exposes correction decisions around symbolic reasoning. Constraint is
normalized row, column, and box inconsistency. The representation is the
perception tensor, provenance execution vector, or correction-aware state.
Recorded exact-grid accuracy is 78.2\% for ISED and 75.6\% for NASR; the
seed-correctness filter, rather than aggregate clean accuracy, defines every
model's eligible fuzzing corpus.

\paragraph{HWF-7.}
Seven-symbol handwritten arithmetic expressions are mutated through same-class
glyph replacement and bounded rendering changes. The expression tree and
numeric target remain unchanged. Grounding measures digit/operator
uncertainty, drift measures the neural or reasoning state, and constraint
measures probability mass assigned to invalid executions. ISED, DOLPHIN, and
NASR receive the same valid candidate stream but expose different raw states.
Clean exact-expression accuracies are 97.75\%, 55.21\%, and 91.08\% for ISED,
DOLPHIN, and NASR, respectively.

\paragraph{ReaSCAN.}
Generated scenes pair layouts with compositional navigation commands.
Photometric changes preserve scene graph and command. NeSyCoCo exposes object
concept scores and the executed concept-composition program. Boundary,
grounding, drift, and novelty are available. No independent constraint checker
is exposed, so constraint is masked for adaptive methods; constraint-only is a
degenerate all-tie diagnostic and is not interpreted as a usable guide.
NeSyCoCo reaches 98.5\% clean test accuracy in this adapter.

\section{Complete Adaptive Results and Weight Analysis}
\label{app:full-results}

This section provides the complete adaptive results and learned \rcw\
allocations. All failure discovery rates are candidate-level means across the
same ten attack seeds used in the main experiments.

\begin{table}[h]
\centering
\scriptsize
\caption{Adaptive improvements over Fixed. $\Delta$ denotes the
percentage-point increase in mean oracle failure discovery rate.}
\label{tab:adaptive-full}
\begin{tabular}{llrrr}
\toprule
Cell & Policy & Fixed & Adaptive & $\Delta$ \\
\midrule
SoC/ISED
    & RCW & 12.675 & 13.832 & +1.157 \\
    & CSG & 12.675 & 13.526 & +0.851 \\

SoC/DOLPHIN
    & RCW & 14.335 & 14.914 & +0.579 \\
    & CSG & 14.335 & 15.557 & +1.222 \\

Sudoku/ISED
    & RCW & 14.133 & 16.062 & +1.929 \\
    & CSG & 14.133 & 16.629 & +2.496 \\

Sudoku/DOLPHIN
    & RCW & 4.908 & 5.764 & +0.856 \\
    & CSG & 4.908 & 5.740 & +0.832 \\

Sudoku/NASR
    & RCW & 21.303 & 23.598 & +2.295 \\
    & CSG & 21.303 & 23.740 & +2.437 \\

HWF/ISED
    & RCW & 6.725 & 9.680 & +2.955 \\
    & CSG & 6.725 & 9.934 & +3.209 \\

HWF/DOLPHIN
    & RCW & 18.098 & 24.719 & +6.621 \\
    & CSG & 18.098 & 23.123 & +5.025 \\

HWF/NASR
    & RCW & 22.385 & 26.650 & +4.265 \\
    & CSG & 22.385 & 25.633 & +3.248 \\

ReaSCAN/NeSyCoCo
    & RCW & 10.211 & 10.283 & +0.072 \\
    & CSG & 10.211 & 10.664 & +0.453 \\
\bottomrule
\end{tabular}
\end{table}

Table~\ref{tab:adaptive-full} gives the complete adaptive improvements over
Fixed. Both \rcw\ and \csg\ improve failure discovery in every evaluated cell,
yielding positive gains in all 18 adaptive comparisons. The largest gains
occur on HWF-7/DOLPHIN, where \rcw\ improves Fixed by $6.621\pp$ and
\csg\ by $5.025\pp$, followed by HWF-7/NASR with gains of $4.265\pp$
and $3.248\pp$, respectively. Substantial improvements also appear on
HWF-7/ISED and Visual Sudoku, while smaller but still positive gains are
observed on Sort-of-CLEVR and ReaSCAN. These results show that adaptive signal
allocation is effective across distinct neural--symbolic interfaces rather than
being driven by a single dataset or model family.

\begin{table}[h]
\centering
\scriptsize
\caption{Learned \rcw\ signal allocations in the order
$(b,g,r,c,n)$. The variation across cells reflects dataset--model-dependent
signal reliability.}
\label{tab:rcw-weights}
\begin{tabular}{lrrrrr}
\toprule
Cell & $b$ & $g$ & $r$ & $c$ & $n$ \\
\midrule
SoC/ISED & .468 & .238 & .228 & .010 & .056 \\
SoC/DOLPHIN & .507 & .261 & .184 & .011 & .036 \\
Sudoku/ISED & .585 & .099 & .083 & .192 & .041 \\
Sudoku/DOLPHIN & .726 & .141 & .021 & .099 & .014 \\
Sudoku/NASR & .623 & .111 & .064 & .175 & .026 \\
HWF/ISED & .446 & .174 & .254 & .090 & .036 \\
HWF/DOLPHIN & .613 & .141 & .184 & .013 & .048 \\
HWF/NASR & .532 & .138 & .277 & .011 & .042 \\
ReaSCAN/NeSyCoCo & .446 & .191 & .255 & .000 & .108 \\
\bottomrule
\end{tabular}
\end{table}

Table~\ref{tab:rcw-weights} shows that the learned \rcw\ allocations vary
substantially across dataset--model cells. Boundary receives the largest weight
in every cell, but its mass ranges from .446 on HWF-7/ISED and
ReaSCAN/NeSyCoCo to .726 on Visual Sudoku/DOLPHIN. Representation drift also
varies strongly, from .021 on Visual Sudoku/DOLPHIN to .277 on HWF-7/NASR,
and receives substantial weight on ReaSCAN/NeSyCoCo (.255). Constraint is
particularly prominent on Visual Sudoku/ISED (.192) and Visual Sudoku/NASR
(.175), but receives little mass in several Sort-of-CLEVR and HWF-7 cells and
is unavailable for ReaSCAN. These differences show that \rcw\ does not merely
recover the same global prior, but reallocates signal mass according to
cell-specific reliability.

The weight patterns also provide a task-level interpretation of the adaptive
results. Visual Sudoku assigns relatively more mass to boundary and constraint
signals, consistent with the interaction between local perceptual errors and
global symbolic constraints. HWF-7 exhibits stronger model dependence:
representation drift receives comparatively high weight for ISED and NASR,
whereas boundary remains dominant for DOLPHIN. Sort-of-CLEVR shows more
moderate adaptive gains despite different learned allocations, while ReaSCAN
demonstrates availability-aware portability: the constraint signal is masked,
yet \csg\ still improves over Fixed using the remaining signals. Detailed
adapter and oracle definitions are provided in
Appendix~\ref{app:adapters}.

\section{Failure-Deduplication Threshold Sensitivity}
\label{app:dedup-sensitivity}

Table~\ref{tab:dedup-threshold} reports the sensitivity of the
activation-similarity clustering results to the cosine-similarity threshold.
Clustering is performed only among failures from the same clean seed and with
the same erroneous structured output; failures with different outputs are
never merged solely based on activation similarity.

\begin{table}[h]
\centering
\scriptsize
\caption{Number of similarity-based failure clusters under different cosine
similarity thresholds.}
\label{tab:dedup-threshold}
\begin{tabular}{c|rrrrrrr}
\toprule
Threshold &
DeepHunter & TensorFuzz & DLFuzz & NSGen-T & CriticalFuzz & \rcw & \csg \\
\midrule
0.95 & 5,400 & 5,910 & 5,610 & 5,540 & 5,490 & 10,982 & 11,659 \\
0.98 & 5,450 & 6,210 & 5,850 & 5,740 & 5,620 & 11,508 & 12,058 \\
0.99 & 5,540 & 6,590 & 6,180 & 5,970 & 5,750 & 12,021 & 12,638 \\
\bottomrule
\end{tabular}
\end{table}

Across all three thresholds, \rcw\ and \csg\ retain substantially more
similarity-based failure clusters than the neural-fuzzing baselines. At
thresholds 0.95, 0.98, and 0.99, the strongest neural baseline retains
5,910, 6,210, and 6,590 clusters, respectively, compared with
10,982, 11,508, and 12,021 for \rcw\ and 11,659, 12,058, and 12,638 for
\csg. The diversity advantage therefore remains consistent across the tested
similarity thresholds.

\section{Wall-Clock Evaluation Protocol}
\label{app:runtime}

We evaluate the wall-clock efficiency of DeepHunter, TensorFuzz, DLFuzz,
NSGen-transfer, CriticalFuzz, Random, Uniform, Fixed, \rcw, and \csg\ using
the same ten paired attack seeds, 20 initially correct inputs, 256 candidate
evaluations per input, beam width eight, mutation families, checkpoints, and
hardware as in the main evaluation. Each timed attack runs alone on one RTX
A6000. Timing includes mutation generation, model execution, feedback or
signal extraction, ranking, and frontier updates, while checkpoint loading
and clean-input validation are excluded. CUDA is synchronized at measured
inference boundaries, and every timed attack executes exactly 256 candidates.

For policy $p$, dataset--model cell $c$, and attack $a$, let
$T^{\mathrm{search}}_{pca}$ denote the total online search time.
We also record the wall-clock time and query index of the first discovered
failure, denoted by $T^{\mathrm{first}}_{pca}$ and
$Q^{\mathrm{first}}_{pca}$, respectively. For attacks that discover no failure,
these quantities are capped at the total search time and 256 queries to avoid
the optimistic bias of averaging only successful attacks.

Failure throughput measures the number of discovered failures per unit online
time:
\begin{equation}
E^{\mathrm{time}}_{pc}=
\frac{\sum_a F_{pca}(T^{\mathrm{search}}_{pca})}
{\sum_a T^{\mathrm{search}}_{pca}},
\end{equation}
where $F_{pca}(t)$ is the cumulative number of oracle failures discovered by
time $t$.

Because absolute inference latency differs substantially across models,
runtime, time to first failure, query count to first failure, and failure
throughput are normalized to \rcw\ within each dataset--model cell. The
resulting multiplicative ratios are aggregated across the nine cells using
the geometric mean. For failure-throughput ratios, cells in which the
reference policy discovers no failure are excluded from the corresponding
geometric mean.

\section{Reproduction Checklist}
\label{app:repro}
The evaluation contains nine compatible cells, twenty policies, and ten
paired repetitions, for $9\times20\times10=1{,}800$ on-policy runs.
Each run samples 20 initially correct inputs and permits 256 candidate
evaluations per valid input with beam width eight. Reproduction proceeds by
(i) verifying checkpoint clean accuracy and the seed-correctness filter;
(ii) fitting \rcw/\csg\ on random-frontier calibration pools without final
outcomes; (iii) executing all policies from scratch under matched inputs,
mutations, validity checks, budgets, and seeds; and (iv) aggregating
candidate-level failure discovery rates by attack seed and reporting their
mean across the ten paired repetitions. A complete audit requires 200
policy--seed summaries per cell and verifies that every policy is present
ten times.

\end{document}